\documentclass{Interspeech2024}

\usepackage{adjustbox}

\interspeechcameraready

\usepackage[table]{xcolor}
\usepackage{tabularx}
\usepackage[capitalise]{cleveref}
\usepackage{hyperref}
\usepackage{breakurl}

\usepackage{multirow}

\usepackage{arydshln} 
\usepackage{booktabs}
\usepackage{multirow}

\usepackage[natbib, bibencoding=utf8, citestyle=numeric, bibstyle=ieee, maxbibnames=999, maxcitenames=2, mincitenames=1, sortcites]{biblatex}
\bibliography{mybib}

\newcommand{\eg}{e.\,g., }

\usepackage[acronym, shortcuts, nohypertypes={acronym}]{glossaries}
\newacronym{AUC}{AUC}{area under the ROC curve}
\newacronym{Bi-LSTM}{Bi-LSTM}{bidirectional long short-term memory neural network}
\newacronym{CCC}{CCC}{concordance correlation coefficient}
\newacronym{BBE}{BBE}{Backbone Block Expansion}
\newacronym{CNN}{CNN}{convolutional neural network}
\newacronym{ECG}{ECG}{electrocardiogram}
\newacronym{GBRT}{GBRT}{Gradient Boosted Regression Trees}
\newacronym{ML}{ML}{machine learning}
\newacronym{MLP}{MLP}{multilayer perceptron}
\newacronym{OM}{OM}{Other Runners Only Model}
\newacronym{RMSE}{RMSE}{root mean squared error}
\newacronym{RPE}{RPE}{received perception of exertion}
\newacronym{SGD}{SGD}{stochastic gradient descent}
\newacronym{sEMG}{sEMG}{surface-electromyography}
\newacronym{UAR}{UAR}{Unweighted Average Recall}
\newacronym{HR}{HR}{heart rate}
\newacronym{PTSD}{PTSD}{Post-traumatic Stress Disorder}
\newacronym{MDD}{MDD}{Major Depressive Disorder}
\newacronym{SVMs}{SVMs}{Support-Vector-Machines}
\newacronym{SVM}{SVM}{Support-Vector-Machine}
\newacronym{TDNN}{TDNN}{Time Delay Neural Network}
\newacronym{NuS}{NuS}{\textit{Der Nordwind und die Sonne} [\textit{The Northwind and the Sun}]}
\newacronym{DtS}{DtS}{\textit{Das tapfere Schneiderlein} [\textit{The Valiant Little Tailor}]}
\newacronym{RBF}{RBF}{radial basis function}
\newacronym{CI}{CI}{confidence interval}
\newacronym{ROC}{ROC}{receiver operating characteristic}
\newacronym{SHAP}{SHAP}{SHapley
Additive exPlanations}
\newacronym{HNR}{HNR}{Harmonics-to-Noise-Ratio}
\newacronym{ITW}{ITW}{\textit{PTSD In The Wild}}
\newacronym{LMU}{LMU}{\textit{LMU PTSD}}
\newacronym{SSL}{SSL}{self-supervised learning}
\newacronym{SER}{SER}{Speech Emotion Recognition}

\title{ExHuBERT: Enhancing HuBERT Through Block Extension and \\Fine-Tuning on 37 Emotion Datasets}

\name[affiliation={1}]{Shahin}{Amiriparian}
\name[affiliation={2}]{Filip}{Packań}
\name[affiliation={2}]{Maurice}{Gerczuk}
\name[affiliation={1,2,3}]{Björn W.}{Schuller}

\address{
  $^1$CHI -- Chair of Health Informatics, MRI, TU Munich, Germany\\
  $^2$Chair of Embedded Intelligence for Health Care \& Wellbeing, University of Augsburg, Germany\\
  $^3$GLAM -- Group on Language, Audio, \& Music, Imperial College, UK
  }
\email{shahin.amiriparian@tum.de}

\keywords{affective computing, speech emotion recognition, transformers, deep learning}

\usepackage{xcolor}

\begin{document}

\makeatletter
\renewcommand\@makefntext[1]{\leftskip=2em\hskip-1em\@makefnmark#1}
\makeatother

\maketitle

\begin{abstract}

    Foundation models have shown great promise in speech emotion recognition (SER) by leveraging their pre-trained representations to capture emotion patterns in speech signals. To further enhance SER performance across various languages and domains, we propose a novel twofold approach. First, we gather EmoSet++, a comprehensive multi-lingual, multi-cultural speech emotion corpus with 37 datasets, 150,907 samples, and a total duration of 119.5 hours. Second, we introduce ExHuBERT, an enhanced version of HuBERT achieved by backbone extension and fine-tuning on \textsc{EmoSet++}. We duplicate each encoder layer and its weights, then freeze the first duplicate, integrating an extra zero-initialized linear layer and skip connections to preserve functionality and ensure its adaptability for subsequent fine-tuning. Our evaluation on unseen datasets shows the efficacy of ExHuBERT, setting a new benchmark for various SER tasks. Model and details on \textsc{EmoSet++}: \mbox{\texttt{\href{https://huggingface.co/amiriparian/ExHuBERT}{https://huggingface.co/amiriparian/ExHuBERT}}}.     
\end{abstract}


\section{Introduction}
\label{sec:intro}


\ac{SER} has a rich research history going back to the 
1970s (first patents)~\cite{williamson1978speech} and 1990s (first research papers)~\cite{dellaert1996recognizing}
and while it has reaped the benefits of deep learning, a core issue remains to this day: While there are many available databases of emotional speech, most of them only contain comparatively few samples or speakers, hindering effective single-corpus training of large neural networks~\cite{Schuller18-SER}. As a response to this circumstance, cross- and multi-corpus \ac{SER} has established itself as a highly important research direction~\cite{Zhang21-DCS}. Due to the fact that databases in the field often differ significantly in recording settings, nature of speech and emotion portrayal (acted, elicited, natural), language, and other factors, successful approaches have employed special strategies and architectural considerations such as domain adaptation~\cite{Latif23-SSA} or adapter transfer learning~\cite{Gerczuk22-EAT} to achieve satisfactory performance. Other efforts have gone towards making deep learning models more robust against distortions, noise, and other variations in the speech signal~\cite{Oates19-RSE,Triantafyllopoulos19-TRS}. 


However, the recent paradigm shift in general deep learning towards large, Transformer-based models has already impacted the field~\cite{Wagner23-DOT}. The Transformer architecture's inherent capability of learning arbitrary structural information from high-dimensional data combined with the exploitation of huge amounts of unlabeled data through self- and unsupervised learning has  significantly reduced the need for human-annotated corpora for training powerful and transferable models~\cite{Vaswani17-AIA,Ericsson22-SRL}. Specifically for the fields of speech recognition and analysis, pre-trained Transformers such as Wav2Vec2.0 (W2V2)~\cite{Baevski20-W2A} or HuBERT~\cite{hsu2021hubert} have shown considerable generalization capabilities. By exploiting self-supervision on large amounts of unlabeled speech data, these models learn to effectively capture the structure of spoken language. For a wide range of downstream tasks, pre-trained transformers provide competitive performance as feature extractors~\cite{Yang21-SSP,Pepino21-ERF} or through finetuning, e.\,g., for \ac{SER} and speaker identification~\cite{Wang22-AFW}. ~\citet{Wagner23-DOT} finetune wav2vec and HuBERT models on MSP-Podcast~\cite{Lotfian19-BNE} and show that their best models provide state-of-the-art performance on a number of \ac{SER} corpora. They further trace the models' efficacy to a number of beneficial characteristics induced by both pre-training and the transformer architecture itself, such as implicit modeling of linguistic information and a general resilience against speaker, gender, or domain variations. 

While these and other works have convincingly made the argument for large, pre-trained Transformer models in \ac{SER}, none have investigated whether their generalisability and robustness can enable effective learning of a single, transferable model on a heterogenous set of databases without the need for domain or corpus adaptation strategies. In the present study, we aim to fill this gap by evaluating the capability of large audio transformers to learn salient features for \ac{SER} by multi-corpus finetuning. For this purpose, we build on the work of~\cite{Gerczuk22-EAT}, introducing EmoSet++, integrating 37 \ac{SER} corpora spanning 15 languages. We then fine-tune HuBERT on the assembled corpus and compare its transfer learning performance to other large pre-trained models on 6 additional \ac{SER} databases. Finally, we introduce ExHuBERT, which integrates \textsc{EmoSet++} fine-tuning with \ac{BBE} -- a technique recently introduced in LLAMA Pro~\cite{Wu24-LPP} -- to deliver a state-of-the-art model and training strategy for emotion recognition. 

\section{EmoSet++}
\label{sec:emoset++}
\begin{figure}[t!]
      \centering
    \includegraphics[width=.8\columnwidth]{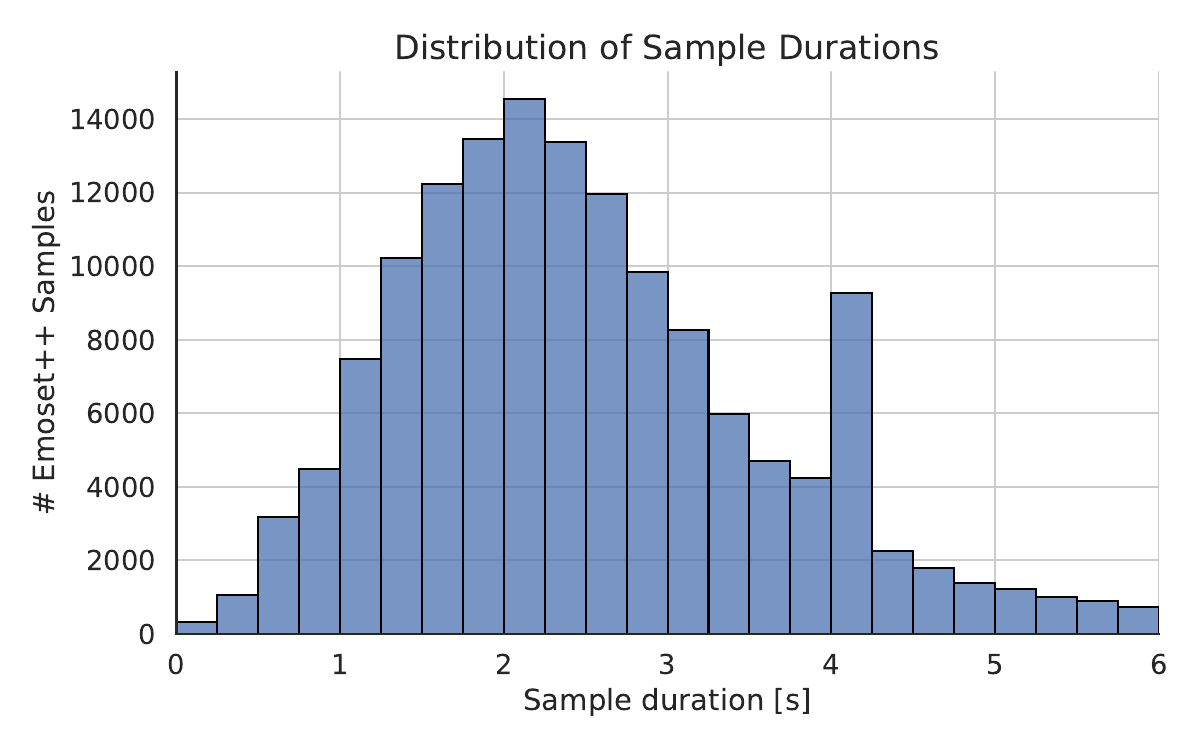}
    \caption{Distribution of speech sample durations within \textsc{EmoSet++}. Predominantly, the samples fall within the range of $0$ to $5$ seconds in length.}
\label{fig:histogram}
\end{figure}

We introduce \textsc{EmoSet++}, a comprehensive multi-lingual, multi-cultural speech emotion corpus. It extends EmoSet~\cite{Gerczuk22-EAT} and integrates $37$ unique emotion datasets with $150\text{,}907$ speech recordings and a cumulative length of $119.5$ hours. Most of the datasets that we have included comprise common languages such as English, German, or Mandarin, alongside rarer ones like Persian or Urdu.
For all corpora, we create speaker-independent splits. To facilitate training, all distinct dataset labels (comprising $106$ emotion classes) are mapped to six classes, representing combinations of low/high arousal and negative/neutral/positive valence. The mapping is based on Russel's Circumplex of Affect~\cite{russell1980circumplex}. 
The dataset splits were derived through three methods: adopting from the collected dataset, manual construction for speaker independence, or simply dividing it into $10\,\%$ partitions for both testing and validation in cases where speaker information was not explicitly provided. We do not conduct any re-weighting or re-balancing of speech samples for our machine learning experiments.~\Cref{fig:histogram} displays sample durations ranging from $0$ to $6$ seconds, with longer samples excluded for readability. Most samples fall in $[0 - 5]$ second range. 

\begin{table*}[htbp]
\centering
\resizebox{1\textwidth}{!}{
\begin{tabular}{lrrrrrrrr}
\toprule
                            &   &   \multicolumn{6}{c}{\textbf{Performance in [\% UAR]}} \\
                            \cmidrule{2-9}

\multicolumn{1}{c}{\textbf{Datasets}} & \multicolumn{1}{c}{W2V2} & \multicolumn{1}{c}{W2V2} & \multicolumn{1}{c}{Whisper} & \multicolumn{1}{c}{Whisper} & \multicolumn{1}{c}{HuBERT} & \multicolumn{1}{c}{HuBERT} & \multicolumn{1}{c}{HuBERT} & \multicolumn{1}{c}{HuBERT} \\


\multicolumn{1}{c}{} & \multicolumn{1}{c}{XLS-R} & \multicolumn{1}{c}{XLS-R} & \multicolumn{1}{c}{Medium} & \multicolumn{1}{c}{Large} & \multicolumn{1}{c}{Large} & \multicolumn{1}{c}{XLarge} & \multicolumn{1}{c}{Large} & \multicolumn{1}{c}{Large} \\

\multicolumn{1}{c}{} & \multicolumn{1}{c}{300M} & \multicolumn{1}{c}{1B} & \multicolumn{1}{c}{} & \multicolumn{1}{c}{v3} & \multicolumn{1}{c}{} & \multicolumn{1}{c}{} & \multicolumn{1}{c}{\textbf{$83.7\,\%$ of \textsc{EmoSet++}}} & \multicolumn{1}{c}{\textbf{\textsc{EmoSet++}}} \\

\midrule    
Berlin Database of Emotional Speech (EMO-DB) \cite{burkhardt2005-DatabaseGermanEmotionala}      & \textbf{91.1} & 86.0 & 55.0 & 61.2 & 90.0 & 82.4 & 88.1 & \cellcolor{gray!40}\textbf{99.1} \\
Database of Elicited Mood in Speech (DEMOS) \cite{parada-cabaleiro2020-DEMoSItalianEmotional}   & 56.1 & 69.2 & 30.0 & 32.8 & \textbf{67.7} & 57.2 & 70.8 & \cellcolor{gray!40}\textbf{90.7} \\
EmoFilm \cite{parada-cabaleiro2018-CategoricalVsDimensional}                                    & 56.9 & 43.4 & 49.7 & 44.5 & \textbf{58.6} & 57.4 & 58.9 & \cellcolor{gray!40}\textbf{60.5} \\
EmotiW-2014\cite{dhall2014-EmotionRecognitionWild}                                              & \textbf{40.0} & 34.7 & 28.6 & 27.2 & 33.1 & 32.5 & \cellcolor{gray!40}\textbf{40.2} & 39.1 \\
eNTERFACE \cite{martin2006-ENTERFACE05AudioVisual}                                              & 66.4 & 76.1 & 39.3 & 38.6 & \textbf{93.9} & 63.2 & 92.9 & \cellcolor{gray!40}\textbf{94.6} \\
IEMOCAP \cite{busso2008-IEMOCAPInteractiveEmotionalb}                                           & 56.4 & 49.7 & 42.9 & 39.0 & 61.1 & \textbf{65.8} & 63.9 & \cellcolor{gray!40}\textbf{67.8} \\
Multimodal EmotionLines Dataset (MELD) \cite{poria2019-MELDMultimodalMultiParty}                & 23.2 & 24.7 & 23.8 & 22.6 & \textbf{30.0} & 25.9 & 34.1 & \cellcolor{gray!40}\textbf{38.5} \\
Mandarin Emotional Speech (MES) \cite{nwe2003-SpeechEmotionRecognition}                         & 66.3 & 48.8 & 66.3 & 65.0 & \textbf{67.5} & 63.8 & \cellcolor{gray!40}\textbf{70.0} & 67.5 \\
\midrule
\textbf{Average UAR over all datasets} & 57.1 & 54.1 & 42.0 & 41.4 & 62.7 & 56.0 & 64.9 & 69.7\\

\bottomrule
\end{tabular}
}
\caption{
Performance comparison of the applied Transformers on eight common speech emotion datasets. Our proposed fine-tuning of HuBERT Large on \textsc{EmoSet++} demonstrates superior performance over all other Transformers. The best results without fine-tuning on \textsc{EmoSet++} are bolded, and the best overall results (including fine-tuning on \textsc{EmoSet++}) are bolded and lightly shaded.
}
\label{tab:dataset_performance}
\end{table*}

\begin{figure*}[h!]
      \centering
    \includegraphics[width=.63\textwidth]{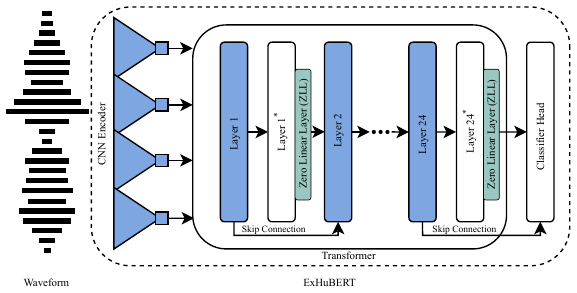}
    \caption{Outline of the proposed ExHuBERT architecture, including skip connections and zero linear layers. The CNN encoder and the weights of the layers colored in blue are frozen. The classifier head and the weights of the layers marked with an asterisk (*) will undergo fine-tuning during the evaluation process.}
\label{fig:sys-digram}
\end{figure*}

\section{Proposed Approach}
\label{sec:system}
We propose an enhanced version of HuBERT~\cite{hsu2021hubert} through fine-tuning on \textsc{EmoSet++} and incorporating Backbone Block Expansion (BBE), denoted as ExHuBERT. Specifically, we fine-tune the encoder part of the HuBERT architecture on $37$ diverse emotion datasets, which span various languages and cultural backgrounds (cf.~\Cref{ssec:hubert-training}). After the weights are updated, we conduct backbone extension (cf.~\Cref{ssec:backbone}), and in the final step, we evaluate the performance of the backbone extended HuBERT (ExHuBERT) on 6 unseen test emotion datasets (cf.~\Cref{sec:exp}).

\subsection{Fine-Tuning HuBERT on EmoSet++}
\label{ssec:hubert-training}
Our system is built upon the Transformer architecture HuBERT~\cite{hsu2021hubert}, which has shown promising results in SER~\cite{pastor2022cross,Wagner23-DOT,wang2021fine}. A simple linear layer is added on top for the classification of the $6$ mapped arousal valence classes. Fine-tuning is conducted in a round-robin fashion, ensuring each dataset contributes equally to the model. Upon achieving a state where our model spans multiple domains and languages, we utilize its transfer learning and generalization capabilities within ExHuBERT.

\subsection{Backbone Block Expansion}
\label{ssec:backbone}
After the fine-tuning process, we duplicate each encoder layer along with its weights. This augmentation results in an expanded version of HuBERT (ExHuBERT), featuring a total of $48$ layers. The newly added layers are inserted after the original ones, accompanied by a skip connection to maintain the original layer's behavior.  To stabilize the training process after layer duplication, we add a Zero Linear Layer (ZLL) at the end of each duplicated layer. The ZLL comprises initialized zero weights, ensuring that the output of the copied layers initiates from zero. This technique plays a crucial role in training by preventing unknown outputs from destabilizing the training process~\cite{Wu24-LPP}. Furthermore, we freeze the original layers to preserve their encoded knowledge, permitting only the copied layers to undergo training. These steps guarantee that the HuBERT model with BBE behaves identically to the HuBERT model without BBE during the initial stages. A high-level overview of ExHuBERT is depicted in~\Cref{fig:sys-digram}.

\section{Experiments and Results}
\label{sec:exp}
We split the experiments into two main parts: i) selection of the suitable audio Transformer for the \ac{BBE} and fine-tuning of the chosen Transformer on \textsc{EmoSet++} (cf.~\Cref{ssec:preselection}), and ii) conducting BBE on the fine-tuned Transformer and evaluating its performance on unseen emotion datasets (cf.~\Cref{ssec:exhubert-test}).

\subsection{Selection of the Suitable Audio Transformer for \ac{BBE}} \label{ssec:preselection}
To choose the optimal architecture for BBE, we evaluate $6$ state-of-the-art Transformers, including W2V2 XLS-R (300 million\footnote{\url{https://huggingface.co/facebook/wav2vec2-xls-r-300m}} and 1 billion\footnote{\url{https://huggingface.co/facebook/wav2vec2-xls-r-1b}}), Whisper (Medium\footnote{\url{https://huggingface.co/openai/whisper-medium}} and Large v3\footnote{\url{https://huggingface.co/openai/whisper-large-v3}}), and HuBERT (Large\footnote{\url{https://huggingface.co/facebook/hubert-large-ls960-ft}} and XLarge\footnote{\url{https://huggingface.co/facebook/hubert-xlarge-ls960-ft}}) on all 26 emotion corpora of \textsc{EmoSet}~\cite{Gerczuk22-EAT}. 
We selected two variants of each architecture to compare their parameter impact, ensuring that variants of the same size had approximately equal parameter counts. Each Transformer is initialized with pre-trained weights obtained from \texttt{huggingface.co}. Additionally, we add a simple linear layer on top of each Transformer, with an output size of $6$, corresponding to the mapped classes. For the evaluation metric, we use \ac{UAR} due to its effectiveness in assessing the overall classification performance across all classes without bias towards any under- or oversampled class. We fine-tune all Transformer models in a round-robin fashion, sequentially passing each dataset forward and backward through the model with one batch in each step. For W2V2 and HuBERT variants, we use raw audio waveforms resampled to $16$ kHz as inputs, while we feed Whisper with log Mel spectrograms (with either $80$ or $128$ bins) obtained from waveforms. We freeze the CNN encoder during the entire training process for W2V2 and HuBERT. This step is unnecessary for Whisper. We conclude the experimentation phase after $3$k steps using AdamW optimisation with $\beta_1 = 0.9$, $\beta_2 = 0.999$, $\epsilon = 1e-08$, and a learning rate of $1e-05$. The performance of these audio Transformers on the unseen test partition of eight commonly used emotion datasets is provided in~\Cref{tab:dataset_performance}\footnote{Results on all datasets are provided here: \\ \url{https://huggingface.co/amiriparian/ExHuBERT/blob/main/supp-mat.pdf}
}. 

Our results demonstrate that the HuBERT Large model outperforms others, achieving an average UAR of $62.7\,\%$ over all eight datasets, followed by W2V2 XLS-R 300 M with $57.1\,\%$ UAR. The worst-performing models are both variants of Whisper, each achieving $42.0\,\%$ and $41.1\,\%$ UAR, respectively. We consequently settle on HuBERT Large as the base model for fine-tuning on the extended \textsc{EmoSet++} and transfer learning through \ac{BBE}.

Subsequently, we test the impact of using both $83.7\,\%$ of \textsc{EmoSet++} and the full \textsc{EmoSet++} for fine-tuning the selected HuBERT model for speech emotion recognition, aiming to evaluate the impact of dataset size on model performance and generalization. Training on  \textsc{EmoSet++} leads to substantial performance gains for all databases, compared to the original \textsc{EmoSet}, raising the average \ac{UAR} over the 8 databases from $62.7\,\%$ to $64.9\,\%$ when only adding 5 training corpora, and to $69.7\,\%$ with the full set. All of the evaluated databases benefit from adding more corpora to \textsc{EmoSet}. However, EmoFilm and Mandarin Emotional Speech stop seeing gains after adding the first 5 new datasets.

\begin{figure*}[h!]
    \centering
    \includegraphics[width=\textwidth]{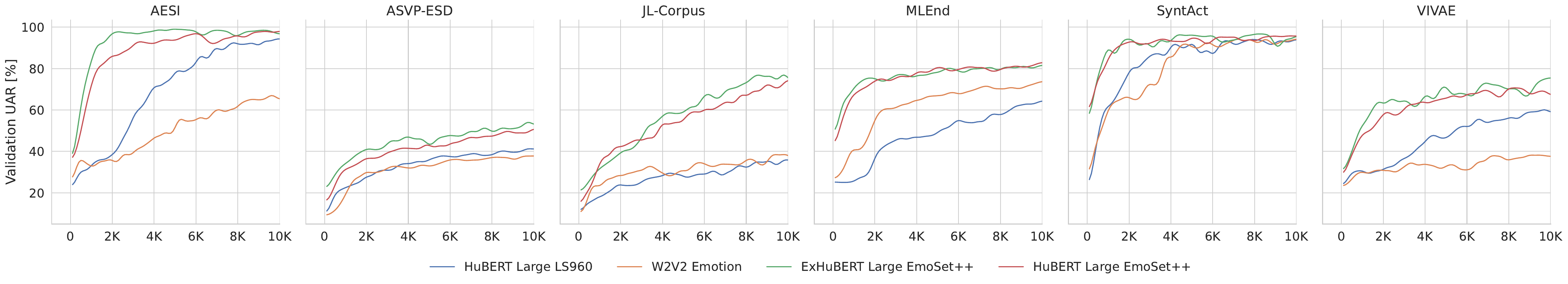}
    \caption{Training curves on the six external \ac{SER} corpora, displaying validation \ac{UAR}.}
    \label{fig:training_plots}
\end{figure*}

\begin{table*}[htbp]
\centering
\resizebox{1\textwidth}{!}{
\begin{tabular}{lrrrrrrrr}
\toprule
                    &   \multicolumn{8}{c}{\textbf{Performance  on unseen test in [\% UAR]}} \\
\cmidrule{2-9}
\textbf{Dataset}    & W2V2  & W2V2  & HuBERT  & HuBERT  & ExHuBERT & ExHuBERT & ExHuBERT & ExHuBERT \\
            & Emotion    & XLS-R  & Large LS960     & Large    & Large LS960     & Large  & Large Non-Frozen & XLarge \\
            &            &       &                 & \textsc{EmoSet++}    &   & \textsc{EmoSet++} & \textsc{EmoSet++} & \textsc{EmoSet++}\\
\midrule
AESI~\cite{chaspari2015-DevelopmentAthensEmotional} & 48.8 &  40.6 & 80.9 & 93.0 & 81.1 & \cellcolor{gray!40}\textbf{94.7} & 86.9 & 33.7 \\
ASVP-ESD~\cite{dejoli2021-AudioSpeechVision} & 36.7 &  34.5 &  40.5 & 45.9 & 41.8 &  \cellcolor{gray!40}\textbf{51.4} &  43.5 &  27.1 \\
JL-Corpus~\cite{james2018-OpenSourceEmotional} &  41.4 & 36.8 & 47.9 & 66.7 & 45.6 & \cellcolor{gray!40}\textbf{67.7} &  47.7 & 41.7 \\
MLEnd & 73.6 & 65.2 & 74.6 & \cellcolor{gray!40}\textbf{81.4} & 77.7 & 80.0 & 77.9 & 60.1 \\
SyntAct~\cite{burkhardt2022-SyntActSynthesizedDatabase} &  82.7 &  89.6 & \cellcolor{gray!40}\textbf{93.7} & 84.6 & 91.2 & 93.0 & 93.5 & 63.9 \\
VIVAE~\cite{holz2022-VariablyIntenseVocalizations} &  45.0 & 45.6 & 42.1 & 54.5 & 46.4 & \cellcolor{gray!40}\textbf{58.2} &  49.3 &  25.1\\
\midrule
\textbf{Average UAR over all datasets} & 52.1 & 54.7 & 63.3 & 71.1 & 61.3 & 74.2 & 66.5 & 42.0 \\
\midrule
\textbf{Number of trainable parameters} &  161.13 M & 311.49 M & 311.49 M & 311.49 M &  336.68 M &  336.68 M &  664.18 M &  664.18 M \\
\bottomrule
\end{tabular}
}
\caption{Performance comparison of the transfer learning capabilities of our proposed backbone extended HuBERT (ExHuBERT), and its variations, against state-of-the-art Transformers across six emotion datasets. None of these datasets were utilized in the fine-tuning process of \textsc{EmoSet++}. For each model, we also provide the average UAR over all datasets and the number of trainable parameters. The best results are bolded and lightly shaded.}
\label{tab:results-test-exhubert}
\end{table*}

\subsection{ExHuBERT Comparison}\label{ssec:exhubert-test}
We compare the transfer learning capabilities of our enhanced version of HuBERT (ExHuBERT Large \textsc{EmoSet++}) against other state-of-the-art Transformer models on a set of six unseen \ac{SER} corpora, including Athens Emotional States Inventory (AESI)~\cite{chaspari2015-DevelopmentAthensEmotional}, Audio, Speech, and Vision Processing Lab Emotional Sound database (ASVP-ESD)~\cite{dejoli2021-AudioSpeechVision}, JL-Corpus~\cite{james2018-OpenSourceEmotional}, MLEnd\footnote{\url{https://mlenddatasets.github.io/spoken_numerals/}}, Synthesized Database of Basic Emotions (SyntAct)~\cite{burkhardt2022-SyntActSynthesizedDatabase}, and Variably Intense Vocalizations of Affect and Emotion Corpus (VIVAE)~\cite{holz2022-VariablyIntenseVocalizations}. Specifically, we chose two variants of W2V2 -- W2V2 XLS-R and the emotion fine-tuned W2V2 model by~\citet{Wagner23-DOT} -- and HuBERT Large LS960 to evaluate the impact of model architectures and the efficacy of \textsc{EmoSet++} fine-tuning. Furthermore, we ablate the performance gains achieved through \textsc{EmoSet++} fine-tuning from those due to block expansion by additionally expanding the LibriSpeech pre-trained HuBERT model (ExHuBERT Large LS960). Finally, we increase the number of trainable parameters of our ExHuBERT model by (1) unfreezing the original HuBERT layers (ExHuBERT Large Non-Frozen \textsc{EmoSet++}) and (2) tripling each original layer during block expansion (ExHuBERT XLarge \textsc{EmoSet++}). \Cref{tab:results-test-exhubert} shows the results and the number of trainable parameters for each of the evaluated models on six databases external to \textsc{EmoSet++}.
For all but one of the six external databases, starting from a model that has been pre-trained on \ac{SER} data, be it MSP-Podcast for W2V2 Emotion or \textsc{EmoSet++} for HuBERT Large, leads to substantially improved performance, compared to the corresponding models trained on general speech data. The noteworthy outlier is found with SyntAct, which is a database of synthesized emotional speech. Here, performance has degraded from the respective base models of W2V2 and HuBERT large, indicating that fine-tuning on emotional speech data might have diminished generalization capabilities to non- and atypical \ac{SER} tasks. However, \ac{BBE} closes this performance gap, enabling efficient transfer from human-recorded to synthetic emotional speech. We hypothesize that without \ac{BBE}, important \ac{SER} knowledge acquired during pre-training gets overwritten early in the fine-tuning process due to the domain gap between synthetic and natural voices. Looking at the remaining databases, \ac{BBE} leads to performance gains for HuBERT fine-tuned on \textsc{EmoSet++}, raising the average \ac{UAR} from $71.1\,\%$ to $74.2\,\%$. On the other hand, applying \ac{BBE} to HuBERT pre-trained on LibriSpeech (ExHuBERT Large LS960) does not lead to consistent improvements across the six \ac{SER} corpora, with a slightly lower average \ac{UAR} of $61.3\,\%$ compared to fine-tuning without \ac{BBE}. Overall, \ac{BBE} seems to have a positive effect on accuracy when the domain shift between source and target databases is rather small, \eg from one \ac{SER} corpus to another. 

We further analyze how fast the different models converge during training on an unseen corpus in \Cref{fig:training_plots}, which shows the validation \acp{UAR} over 10,000 training steps. Analogous to the test set results, \textsc{EmoSet++} pre-training increases overall performance and further achieves faster convergence on all databases compared to models without \ac{SER} pre-training and the MSP-Podcast fine-tuned W2V2. For AESI, \ac{BBE} helps the ExHuBERT model to converge even earlier.  

To conclude, we look at the results achieved with the larger versions of ExHuBERT, which double the amount of trainable parameters. Unfreezing the original HuBERT layers after \ac{BBE} degrades performance even from simple fine-tuning of HuBERT on each of the target corpora, highlighting the importance of keeping the weights of the original model fixed for transfer learning. Expanding each block in ExHuBERT by adding a second copy of the respective layer suffers from substantial overfitting, achieving the worst overall \ac{UAR} of all evaluated Transformer models.

In summary, the validation of our approach on both databases contained within \textsc{EmoSet++} and external corpora shows that (1) multi-corpus pre-training on \textsc{EmoSet++} leads to substantially increased \ac{SER} performance on seen and unseen corpora, and (2) the addition \ac{BBE} in ExHuBERT further helps with generalization to new datasets.

For running all of our machine learning experiments, we utilized one RTX-3090 GPU with $24$\,GB memory, and needed a total time of $313$\,hours: $121$\,h for stage 1 (pre-selection of Transformers), $87$\,h for \textsc{EmoSet++} fine-tuning, $17$\,h for ExHuBERT \textsc{EmoSet++} testing, and $88$\,h for the second stage.

\section{Conclusions}
We have proposed a novel twofold approach for \ac{SER} by (i) collecting \textsc{EmoSet++}, a comprehensive multi-cultural and multi-lingual corpus comprising 37 emotion datasets, and (ii) introducing an enhanced version of HuBERT, denoted as ExHuBERT, achieved through fine-tuning on \textsc{EmoSet++} and incorporating backbone extension. To find the suitable Transformer for BBE, we first selected six versions of state-of-the-art audio Transformers and analyzed their performance on \textsc{EmoSet}~\cite{Gerczuk22-EAT} and then fine-tuned the best-performing Transformer (which was HuBERT Large) on \textsc{EmoSet++}.  In the subsequent phase, we applied BBE to the fine-tuned HuBERT Large model and compared its performance with other Transformers on six previously unseen emotion datasets. The experimental results underscore the effectiveness of our proposed approach, demonstrating its capacity to generalize across diverse datasets and establish new benchmarks for a variety of emotion recognition tasks. Lastly, we have uploaded ExHuBERT on \texttt{huggingface.co}\footnote{
\url{https://huggingface.co/amiriparian/ExHuBERT}}. 
Fine-tuning and deploying ExHuBERT may be computationally demanding, potentially \textbf{limiting} its use in resource-constrained environments. 

For future work, we aim to include MSP-Podcast dataset~\cite{Lotfian19-BNE} in \textsc{EmoSet++} and enhance ExHuBERT for continuous recognition of arousal, valence, and dominance.

\section{Acknowledgements}
This work was supported by MDSI -- Munich Data Science Institute as well as MCML -- Munich Center of Machine Learning.

\clearpage\newpage
\section{\refname}
\printbibliography[heading=none]

@inproceedings{dellaert1996recognizing,
  title={Recognizing emotion in speech},
  author={Dellaert, Frank and Polzin, Thomas and Waibel, Alex},
  booktitle={Proc. ICSLP},
  volume={3},
  pages={1970--1973},
  year={1996},
  organization={IEEE}
}

@misc{williamson1978speech,
  title={Speech analyzer for analyzing pitch or frequency perturbations in individual speech pattern to determine the emotional state of the person},
  author={Williamson, John Decatur},
  year={1978},
  month=jun # "~6",
  publisher={Google Patents},
  note={US Patent 4,093,821}
}

@article{wang2021fine,
  title={A fine-tuned wav2vec 2.0/hubert benchmark for speech emotion recognition, speaker verification and spoken language understanding},
  author={Wang, Yingzhi and Boumadane, Abdelmoumene and Heba, Abdelwahab},
  journal={arXiv preprint arXiv:2111.02735},
  year={2021}
}

@inproceedings{pastor2022cross,
  title={Cross-corpus speech emotion recognition with HuBERT self-supervised representation},
  author={Pastor, Miguel A and Ribas, Dayana and Ortega, Alfonso and Miguel, Antonio and Lleida, Eduardo},
  booktitle={IberSPEECH},
  pages={76--80},
  year={2022},
  organization={ISCA}
}

@article{hsu2021hubert,
  title={Hubert: Self-supervised speech representation learning by masked prediction of hidden units},
  author={Hsu, Wei-Ning and Bolte, Benjamin and Tsai, Yao-Hung Hubert and Lakhotia, Kushal and Salakhutdinov, Ruslan and Mohamed, Abdelrahman},
  journal={IEEE/ACM Transactions on Audio, Speech, and Language Processing},
  volume={29},
  pages={3451--3460},
  year={2021},
  publisher={IEEE}
}

@article{russell1980circumplex,
  title={A circumplex model of affect.},
  author={Russell, James A},
  journal={Journal of personality and social psychology},
  volume={39},
  number={6},
  pages={1161},
  year={1980},
  publisher={American Psychological Association}
}

@article{Wagner23-DOT,
  title = {Dawn of the Transformer Era in Speech Emotion Recognition: Closing the Valence Gap},
  author = {Wagner, Johannes and Triantafyllopoulos, Andreas and Wierstorf, Hagen and Schmitt, Maximilian and Burkhardt, Felix and Eyben, Florian and Schuller, Bj{\"o}rn W.},
  year = {2023},
  month = sep,
  journal = {IEEE Transactions on Pattern Analysis and Machine Intelligence},
  volume = {45},
  number = {9},
  pages = {10745--10759},
  %%issn = {0162-8828, 2160-9292, 1939-3539},
  %%doi = {10.1109/tpami.2023.3263585},
  langid = {american}
}

@misc{Wang22-AFW,
  title = {A {{Fine-tuned Wav2vec}} 2.0/{{HuBERT Benchmark For Speech Emotion Recognition}}, {{Speaker Verification}} and {{Spoken Language Understanding}}},
  author = {Wang, Yingzhi and Boumadane, Abdelmoumene and Heba, Abdelwahab},
  year = {2022},
  month = oct,
  number = {arXiv:2111.02735},
  eprint = {2111.02735},
  primaryclass = {cs, eess},
  publisher = {{arXiv}},
  archiveprefix = {arxiv},
  langid = {english}
}

@inproceedings{Yang21-SSP,
  title = {{{SUPERB}}: Speech Processing Universal {{PERformance}} Benchmark},
  shorttitle = {{{SUPERB}}},
  booktitle = {Proc. INTERSPEECH},
  author = {Yang, Shu-wen and Chi, Po-Han and Chuang, Yung-Sung and others},
  year = {2021},
  month = aug,
  pages = {1194--1198},
  publisher = {{ISCA}},
  %%doi = {10.21437/Interspeech.2021-1775},
  langid = {american}
}

@article{Pepino21-ERF,
  title = {Emotion Recognition from Speech Using Wav2vec 2.0 Embeddings},
  author = {Pepino, Leonardo and Riera, Pablo and Ferrer, Luciana},
  year = {2021},
  month = aug,
  journal = {Interspeech 2021},
  pages = {3400--3404},
  publisher = {{ISCA}},
  %%doi = {10.21437/Interspeech.2021-703},
  langid = {american}
}

@article{Latif23-SSA,
  title = {Self Supervised Adversarial Domain Adaptation for Cross-Corpus and Cross-Language Speech Emotion Recognition},
  author = {Latif, Siddique and Rana, Rajib and Khalifa, Sara and Jurdak, Raja and Schuller, Bj{\"o}rn},
  year = {2023},
  month = jul,
  journal = {IEEE Transactions on Affective Computing},
  volume = {14},
  number = {3},
  pages = {1912--1926},
  %%issn = {1949-3045, 2371-9850},
  %%doi = {10.1109/TAFFC.2022.3167013},
  langid = {american}
}

@misc{Wu24-LPP,
  title = {{{LLaMA}} pro: Progressive {{LLaMA}} with Block Expansion},
  shorttitle = {{{LLaMA}} Pro},
  author = {Wu, Chengyue and Gan, Yukang and Ge, Yixiao and Lu, Zeyu and Wang, Jiahao and Feng, Ye and Luo, Ping and Shan, Ying},
  year = {2024},
  month = jan,
  number = {arXiv:2401.02415},
  eprint = {2401.02415},
  primaryclass = {cs},
  publisher = {{arXiv}},
  archiveprefix = {arxiv},
  langid = {english}
}

@article{Schuller18-SER,
  title = {Speech Emotion Recognition: Two Decades in a Nutshell, Benchmarks, and Ongoing Trends},
  shorttitle = {Speech Emotion Recognition},
  author = {Schuller, Bj{\"o}rn W.},
  year = {2018},
  month = apr,
  journal = {Communications of the ACM},
  volume = {61},
  number = {5},
  pages = {90--99},
%  %issn = {0001-0782, 1557-7317},
%  %doi = {10.1145/3129340},
  langid = {american}
}

@inproceedings{Baevski20-W2A,
  title = {Wav2vec 2.0: A Framework for Self-Supervised Learning of Speech Representations},
  shorttitle = {Wav2vec 2.0},
  booktitle = {Proc. NeurIPS},
  author = {Baevski, Alexei and Zhou, Yuhao and Mohamed, Abdelrahman and Auli, Michael},
  year = {2020},
  volume = {33},
  pages = {12449--12460},
  publisher = {{Curran Associates, Inc.}}
}

@article{Lotfian19-BNE,
  title = {Building Naturalistic Emotionally Balanced Speech Corpus by Retrieving Emotional Speech from Existing Podcast Recordings},
  author = {Lotfian, Reza and Busso, Carlos},
  year = {2019},
  month = oct,
  journal = {IEEE Transactions on Affective Computing},
  volume = {10},
  number = {4},
  pages = {471--483},
  %issn = {1949-3045, 2371-9850},
  %%doi = {10.1109/TAFFC.2017.2736999},
  langid = {american}
}

@article{Ericsson22-SRL,
  title = {Self-Supervised Representation Learning: Introduction, Advances, and Challenges},
  shorttitle = {Self-Supervised Representation Learning},
  author = {Ericsson, Linus and Gouk, Henry and Loy, Chen Change and Hospedales, Timothy M.},
  year = {2022},
  month = may,
  journal = {IEEE Signal Processing Magazine},
  volume = {39},
  number = {3},
  pages = {42--62},
  %%issn = {1053-5888, 1558-0792},
  %%doi = {10.1109/MSP.2021.3134634},
  langid = {american},
  keywords = {Annotations,Computational efficiency,Deep learning,Representation learning,Self-supervised learning}
}

@inproceedings{Vaswani17-AIA,
  title = {Attention Is All You Need},
  booktitle = {Proc. NeurIPS},
  author = {Vaswani, Ashish and Shazeer, Noam and Parmar, Niki and Uszkoreit, Jakob and Jones, Llion and Gomez, Aidan N and Kaiser, {\L}ukasz and Polosukhin, Illia},
  editor = {Guyon, I. and Luxburg, U. Von and Bengio, S. and Wallach, H. and Fergus, R. and Vishwanathan, S. and Garnett, R.},
  year = {2017},
  volume = {30},
  publisher = {{Curran Associates, Inc.}},
  langid = {american}
}

@inproceedings{Oates19-RSE,
  title = {Robust Speech Emotion Recognition under Different Encoding Conditions},
  booktitle = {Proc. INTERSPEECH},
  author = {Oates, Christopher and Triantafyllopoulos, Andreas and Steiner, Ingmar and Schuller, Bj{\"o}rn W.},
  year = {2019},
  month = sep,
  pages = {3935--3939},
  publisher = {{ISCA}},
  %%doi = {10.21437/Interspeech.2019-1658},
  langid = {american}
}

@inproceedings{Triantafyllopoulos19-TRS,
  title = {Towards Robust Speech Emotion Recognition Using Deep Residual Networks for Speech Enhancement},
  booktitle = {Proc. INTERSPEECH},
  author = {Triantafyllopoulos, Andreas and Keren, Gil and Wagner, Johannes and Steiner, Ingmar and Schuller, Bj{\"o}rn W.},
  year = {2019},
  month = sep,
  pages = {1691--1695},
  publisher = {{ISCA}},
  %%doi = {10.21437/Interspeech.2019-1811},
  langid = {american}
}

@article{Zhang21-DCS,
  title = {Deep Cross-Corpus Speech Emotion Recognition: Recent Advances and Perspectives},
  shorttitle = {Deep Cross-Corpus Speech Emotion Recognition},
  author = {Zhang, Shiqing and Liu, Ruixin and Tao, Xin and Zhao, Xiaoming},
  year = {2021},
  journal = {Frontiers in Neurorobotics},
  volume = {15},
  % %issn = {1662-5218},
  % %doi = {10.3389/fnbot.2021.784514},
  langid = {american}
}

@article{Gerczuk22-EAT,
    author = {Maurice Gerczuk and Shahin Amiriparian and Sandra Ottl and Bj\"orn Schuller},
    title = {{EmoNet: A Transfer Learning Framework for Multi-Corpus Speech Emotion Recognition}},
	journal = {IEEE Transactions on Affective Computing},
	year = {2022},
	volume = {13},
	number = {},
	pages = {},
	month = {},
	publisher = {IEEE},
}

@article{chaspari2015-DevelopmentAthensEmotional,
  title = {The Development of the {{Athens Emotional States Inventory}} ({{AESI}}): Collection, Validation and Automatic Processing of Emotionally Loaded Sentences},
  shorttitle = {The Development of the {{Athens Emotional States Inventory}} ({{AESI}})},
  author = {Chaspari, Theodora and Soldatos, Constantin and Maragos, Petros},
  year = {2015},
  journal = {The World Journal of Biological Psychiatry: The Official Journal of the World Federation of Societies of Biological Psychiatry},
  volume = {16},
  number = {5},
  pages = {312--322},
  %issn = {1814-1412},
  %doi = {10.3109/15622975.2015.1012228},
  abstract = {OBJECTIVES: The development of ecologically valid procedures for collecting reliable and unbiased emotional data towards computer interfaces with social and affective intelligence targeting patients with mental disorders. METHODS: Following its development, presented with, the Athens Emotional States Inventory (AESI) proposes the design, recording and validation of an audiovisual database for five emotional states: anger, fear, joy, sadness and neutral. The items of the AESI consist of sentences each having content indicative of the corresponding emotion. Emotional content was assessed through a survey of 40 young participants with a questionnaire following the Latin square design. The emotional sentences that were correctly identified by 85\% of the participants were recorded in a soundproof room with microphones and cameras. A preliminary validation of AESI is performed through automatic emotion recognition experiments from speech. RESULTS: The resulting database contains 696 recorded utterances in Greek language by 20 native speakers and has a total duration of approximately 28 min. Speech classification results yield accuracy up to 75.15\% for automatically recognizing the emotions in AESI. CONCLUSIONS: These results indicate the usefulness of our approach for collecting emotional data with reliable content, balanced across classes and with reduced environmental variability.},
  langid = {english},
  pmid = {25797829},
  keywords = {Adult,automatic data processing,data collection,Data Collection,{Databases, Factual},Emotions,expressed emotion,factual databases,Greece,Humans,Language,Natural Language Processing,questionnaires,Young Adult}
}

@article{dejoli2021-AudioSpeechVision,
  title = {Audio,{{Speech}} and {{Vision Processing Lab Emotional Sound}} Database ({{ASVP-ESD}})},
  author = {Dejoli, Tientcheu Touko Landry and He, Qianhua and Xie, Wei},
  year = {2021},
  month = may,
  %doi = {10.5281/zenodo.4782712},
  %urldate = {2024-02-13},
  abstract = {The Audio, Speech, and Vision Processing Lab Emotional~Sound database~(ASVP-ESD) Dejoli Tientcheu Touko Landry; Qianhua He; Wei Xie Citing the ASVP-ESD The ASVP-ESD emotional sound database is released by Audio, Speech, and Vision Processing Lab(http://www.speech-led.com/main.htm, from the South China University of Technology), so please cite the ASVP-ESD if it is used in your work in any form. ~Personal works, such as machine learning projects or posts, should provide a URL to this Zenodo page, though a reference.~ Contact Information If you would like further information about the ASVP-ESD, when~facing any issues downloading files, please contact us at~~201722800077@mail.scut.edu.cn,~1197581424@qq.com data labeling process The first version~dataset labeling~ (containing 52 folders) was done by 5 different annotators through a tagging application specially designed~for audio tagging the latest~added folders~were done by 3 others annotators.~After listening to each audio the judge should choose the corresponding label according to personal feeling.~Then after the tagging part a simple voting algorithm was build for voting and upgrading the corresponding audio to the class having the most number of vote. Construction and Validation The ASVP-ESD~contains 7812~audio files(with additional 1204 files for babies' voices ).~It is an emotional-based database, containing speech and non-speech emotional sound. ~~The audio were recorded and collected from movies, tv show, youtube channel and others emotional sound website. Comparing to other public emotional databases, ASVP-ESD is more realistic and non-scripted with no language restriction. .Description The Audio, Speech, and Vision Processing Lab Emotional Sound database(ASVP-ESD) contains 7812~ audio files regrouped in 78 folders.~~The data are organized as follows: ~odd folder number for female, even for male~(total size: 1.26 GB). As it's a realistic dataset some folders contain dialog or several people interacting in the audio.~Speech and non-speech Emotional sound include boredom, neutral, happiness (laugh, gaggle), sadness(cry, sniff), angry, fear (scream, deep breath, panic), surprise(amazed), disgust, excite(agitation), pleasure, pain, disappointment~expressions total of 12 different emotions.~ 2 levels of intensity were used for the database (normal and high). Audio is available in 16k, 1 channel, .wav format, the average length of the file is between 0.5 to 20 seconds, for a total of about 10 hours 51 minutes. Note, there are two additional folders (acteur\_150,acteur\_50) that contains only babies audio(laugh, cry)~Audio-only files are regrouped as: Actor\_00 ~is composed of mixed audio samples from movies and website sounds. From actor\_01 to actor\_19 and actor\_31 to acteur\_38 ~are different actors from 3 different movie sounds. From actor\_21 to actor\_29 ~and ~actor\_39 to actor\_68~are only for sounds randomly collected on online platform. Actor\_100(actor\_100 are crowd or many people voices) ~to actor\_102 are from the same website ,same for actor\_103 to actor\_106~ File naming convention Each of the audio files has a unique filename. The filename consists of ~numerical identifier (e.g., 02-01-06-01-02-105-02-01-02.wav, for ~speech /~02-01-06-01-02-105-02-01-02-01-03.wav, for non-speech)~These identifiers define the stimulus characteristics:~ Filename identifiers~ Modality ( 03 = audio-only). Vocal channel (01 = speech, 02 = non speech). Emotion ( 01 = boredom, 02 = neutral, 03 = happy, 04 = sad, 05 = angry, 06 = fearful, 07 = disgust, 08 = surprised, 09 = excited, 10 = pleasure, 11 = pain, 12 = disappointment). Emotional intensity (01 = normal, 02 = high). Statement (as it's non scripted this refer to the number of sample select per actor folder ). Actor ( even~numbered acteurs are male, odd~numbered actors are female). Age~(01 = above 65, 02 = between 20{\textasciitilde}64, 03 = under 20,04=new born). Source of downloading (01 =website , 02 = youtube channel, 03= movies). Language(01=Chinese , 02=English ,04 = french , others)~ Filename example: 03-01-06-01-02-12-02-01-01-16-04.wav: 1.audio-only (03) 2.Speech (01) 3.Fearful (06) 4.Normal intensity (01) 5.Statement ~(02) 6.12th Actorr (12)~folder 12 male as its even 7.Age(02) 8.Source(01) 9.language(01) 10.Screaming ``only for non speech''~(16) 11.the 4th~sample from the same dialog(04) All file with 77 at the end means file with a high noise environment. for non-speech data: Happyness~ is a collection of (laugh=13,gaggle=23,others=33) sadness is a collection of (cry=14, sigh=24,sniffle=34,suffering=44) fear is a collection of (scream 16, breath=26 ,panic=36) angry (rage=15,frustration=25 ,other=35) surprise (surprised=18, amazed=28 ,astonishment=38,others=48) disgust(disgust=17, rejection=27) For any suggestion please don't~ hesitate just send us an email.},
  keywords = {emotional sound,{Speech ,non-speech}}
}

@inproceedings{james2018-OpenSourceEmotional,
  title = {An {{Open Source Emotional Speech Corpus}} for {{Human Robot Interaction Applications}}},
  booktitle = {Proc. INTERSPEECH},
  author = {James, Jesin and Tian, Li and Inez Watson, Catherine},
  year = {2018},
  pages = {2768--2772},
  %doi = {10.21437/Interspeech.2018-1349},
  %urldate = {2024-02-13}
}

@inproceedings{burkhardt2022-SyntActSynthesizedDatabase,
  title = {{{SyntAct}}: {{A Synthesized Database}} of {{Basic Emotions}}},
  shorttitle = {{{SyntAct}}},
  booktitle = {Proc. DCLRL},
  author = {Burkhardt, Felix and Eyben, Florian and Schuller, Bj{\"o}rn},
  editor = {S{\"a}lev{\"a}, Jonne and Lignos, Constantine},
  year = {2022},
  month = jun,
  pages = {1--9},
  publisher = {{European Language Resources Association}},
  address = {{Marseille, France}},
  %urldate = {2024-02-13},
  abstract = {Speech emotion recognition is in the focus of research since several decades and has many applications. One problem is sparse data for supervised learning. One way to tackle this problem is the synthesis of data with emotion simulating speech synthesis approaches. We present a synthesized database of five basic emotions and neutral expression based on rule based manipulation for a diphone synthesizer which we release to the public. The database has been validated in several machine learning experiments as a training set to detect emotional expression from natural speech data. The scripts to generate such a database have been made open source and could be used to aid speech emotion recognition for a low resourced language, as MBROLA supports 35 languages},
  file = {C:\Users\Filip\Zotero\storage\PWIQBC8D\Burkhardt et al. - 2022 - SyntAct A Synthesized Database of Basic Emotions.pdf}
}

@article{holz2022-VariablyIntenseVocalizations,
  title = {The Variably Intense Vocalizations of Affect and Emotion ({{VIVAE}}) Corpus Prompts New Perspective on Nonspeech Perception},
  author = {Holz, Natalie and {Larrouy-Maestri}, Pauline and Poeppel, David},
  year = {2022},
  journal = {Emotion},
  volume = {22},
  number = {1},
  pages = {213--225},
  publisher = {{American Psychological Association}},
  address = {{US}},
  %issn = {1931-1516},
  %doi = {10.1037/emo0001048},
  abstract = {The human voice is a potent source of information to signal emotion. Nonspeech vocalizations (e.g., laughter, crying, moans, or screams), in particular, can elicit compelling affective experiences. Consensus exists that the emotional intensity of such expressions matters; however how intensity affects such signals, and their perception remains controversial and poorly understood. One reason is the lack of appropriate data sets. We have developed a comprehensive stimulus set of nonverbal vocalizations, the first corpus to represent emotion intensity from one extreme to the other, in order to resolve the empirically underdetermined basis of emotion intensity. The full set, comprising 1085 stimuli, features eleven speakers expressing 3 positive (achievement/triumph, sexual pleasure, surprise) and 3 negative (anger, fear, physical pain) affective states, each varying from low to peak emotion intensity. The smaller core set of 480 files represents a fully crossed subsample (6 emotions {\texttimes} 4 intensities {\texttimes} 10 speakers {\texttimes} 2 items) selected based on judged authenticity. Perceptual validation and acoustic characterization of the stimuli are provided; the expressed emotional intensity, like expressed emotion, is reflected in listener evaluation and signal properties of nonverbal vocalizations. These carefully curated new materials can help disambiguate foundational questions on the communication of affect and emotion in the psychological and neural sciences and strengthen our theoretical understanding of this domain of emotional experience. (PsycInfo Database Record (c) 2022 APA, all rights reserved)},
  keywords = {Emotion Recognition,Emotions,Negative Emotions,Positive Emotions,Speech Perception,Stimulus Intensity,Vocalization,Voice},
  file = {C:\Users\Filip\Zotero\storage\WFQ5CHUC\doiLanding.html}
}

@misc{-ChineseLDCOrgZiYuanChaKan,
  title = {- {{ChineseLDC}}.{{Org}} -},
  %urldate = {2024-02-13},
  howpublished = {http://www.chineseldc.org/resource\_info.php?rid=76}
}

@article{banse1996-AcousticProfilesVocal,
  title = {Acoustic Profiles in Vocal Emotion Expression},
  author = {Banse, Rainer and Scherer, Klaus R.},
  year = {1996},
  journal = {Journal of Personality and Social Psychology},
  volume = {70},
  number = {3},
  pages = {614--636},
  publisher = {American Psychological Association},
  address = {US},
  %issn = {1939-1315},
  %doi = {10.1037/0022-3514.70.3.614},
  abstract = {Professional actors' portrayals of 14 emotions varying in intensity and valence were presented to judges. The results on decoding replicated earlier findings on the ability of judges to infer vocally expressed emotions with much-better-than-chance accuracy, including consistently found differences in the recognizability of different emotions. A total of 224 portrayals were subjected to digital acoustical analysis to obtain profiles of vocal parameters for different emotions. The data suggest that vocal parameters not only index the degree of intensity typical for different emotions but also differentiate valence or quality aspects. The data are also used to test theoretical predictions on vocal patterning based on the component process of model of emotion (K. R. Scherer, see record 1986-16849-001). Although most hypotheses are supported, some need to be revised on the basis of the empirical evidence. Discriminant analysis and jackknifing show remarkably high hit rates and patterns of confusion that closely mirror those found for listener-judges. (PsycINFO Database Record (c) 2016 APA, all rights reserved)},
  keywords = {Emotional States,Inference,Oral Communication,Speech Characteristics,Speech Perception}
}

@article{batliner2008-ReleasingThoroughlyAnnotated,
  title = {Releasing a Thoroughly Annotated and Processed Spontaneous Emotional Database: The {{FAU Aibo Emotion Corpus}}},
  author = {Batliner, A and Steidl, S and Noth, E},
  year = {2008},
  langid = {english}
}

@inproceedings{burkhardt2005-DatabaseGermanEmotionala,
  title = {A Database of {{German}} Emotional Speech},
  booktitle = {Proc. INTERSPEECH},
  author = {Burkhardt, Felix and Paeschke, A. and Rolfes, M. and Sendlmeier, Walter F. and Weiss, Benjamin},
  year = {2005},
  month = sep,
  pages = {1517--1520},
  publisher = {ISCA},
  %%doi = {10.21437/Interspeech.2005-446},
  %urldate = {2024-02-13},
  abstract = {The article describes a database of emotional speech. Ten actors (5 female and 5 male) simulated the emotions, producing 10 German utterances (5 short and 5 longer sentences) which could be used in everyday communication and are interpretable in all applied emotions.},
  langid = {english}
}

@article{busso2008-IEMOCAPInteractiveEmotionalb,
  title = {{{IEMOCAP}}: Interactive Emotional Dyadic Motion Capture Database},
  shorttitle = {{{IEMOCAP}}},
  author = {Busso, Carlos and Bulut, Murtaza and Lee, Chi-Chun and Kazemzadeh, Abe and Mower, Emily and Kim, Samuel and Chang, Jeannette N. and Lee, Sungbok and Narayanan, Shrikanth S.},
  year = {2008},
  month = dec,
  journal = {Language Resources and Evaluation},
  volume = {42},
  number = {4},
  pages = {335--359},
  %issn = {1574-0218},
  %%doi = {10.1007/s10579-008-9076-6},
  %urldate = {2024-02-13},
  abstract = {Since emotions are expressed through a combination of verbal and non-verbal channels, a joint analysis of speech and gestures is required to understand expressive human communication. To facilitate such investigations, this paper describes a new corpus named the ``interactive emotional dyadic motion capture database'' (IEMOCAP), collected by the Speech Analysis and Interpretation Laboratory (SAIL) at the University of Southern California (USC). This database was recorded from ten actors in dyadic sessions with markers on the face, head, and hands, which provide detailed information about their facial expressions and hand movements during scripted and spontaneous spoken communication scenarios. The actors performed selected emotional scripts and also improvised hypothetical scenarios designed to elicit specific types of emotions (happiness, anger, sadness, frustration and neutral state). The corpus contains approximately 12~h of data. The detailed motion capture information, the interactive setting to elicit authentic emotions, and the size of the database make this corpus a valuable addition to the existing databases in the community for the study and modeling of multimodal and expressive human communication.},
  langid = {english},
  keywords = {Audio-visual database,Dyadic interaction,Emotion,Emotional assessment,Motion capture system}
}

@article{cao2014-CREMADCrowdsourcedEmotionala,
  title = {{{CREMA-D}}: {{Crowd-sourced Emotional Multimodal Actors Dataset}}},
  shorttitle = {{{CREMA-D}}},
  author = {Cao, Houwei and Cooper, David G. and Keutmann, Michael K. and Gur, Ruben C. and Nenkova, Ani and Verma, Ragini},
  year = {2014},
  journal = {IEEE transactions on affective computing},
  volume = {5},
  number = {4},
  pages = {377--390},
  %issn = {1949-3045},
  %%doi = {10.1109/TAFFC.2014.2336244},
  %urldate = {2024-02-13},
  abstract = {People convey their emotional state in their face and voice. We present an audio-visual data set uniquely suited for the study of multi-modal emotion expression and perception. The data set consists of facial and vocal emotional expressions in sentences spoken in a range of basic emotional states (happy, sad, anger, fear, disgust, and neutral). 7,442 clips of 91 actors with diverse ethnic backgrounds were rated by multiple raters in three modalities: audio, visual, and audio-visual. Categorical emotion labels and real-value intensity values for the perceived emotion were collected using crowd-sourcing from 2,443 raters. The human recognition of intended emotion for the audio-only, visual-only, and audio-visual data are 40.9\%, 58.2\% and 63.6\% respectively. Recognition rates are highest for neutral, followed by happy, anger, disgust, fear, and sad. Average intensity levels of emotion are rated highest for visual-only perception. The accurate recognition of disgust and fear requires simultaneous audio-visual cues, while anger and happiness can be well recognized based on evidence from a single modality. The large dataset we introduce can be used to probe other questions concerning the audio-visual perception of emotion.},
  pmcid = {PMC4313618},
  pmid = {25653738}
}

@inproceedings{costantini2014-EMOVOCorpusItalian,
  title = {{{EMOVO Corpus}}: An {{Italian Emotional Speech Database}}},
  shorttitle = {{{EMOVO Corpus}}},
  booktitle = {Proceedings of the {{Ninth International Conference}} on {{Language Resources}} and {{Evaluation}} ({{LREC}}'14)},
  author = {Costantini, Giovanni and Iaderola, Iacopo and Paoloni, Andrea and Todisco, Massimiliano},
  editor = {Calzolari, Nicoletta and Choukri, Khalid and Declerck, Thierry and Loftsson, Hrafn and Maegaard, Bente and Mariani, Joseph and Moreno, Asuncion and Odijk, Jan and Piperidis, Stelios},
  year = {2014},
  month = may,
  pages = {3501--3504},
  publisher = {European Language Resources Association (ELRA)},
  address = {Reykjavik, Iceland},
  %urldate = {2024-02-13},
  abstract = {This article describes the first emotional corpus, named EMOVO, applicable to Italian language,. It is a database built from the voices of up to 6 actors who played 14 sentences simulating 6 emotional states (disgust, fear, anger, joy, surprise, sadness) plus the neutral state. These emotions are the well-known Big Six found in most of the literature related to emotional speech. The recordings were made with professional equipment in the Fondazione Ugo Bordoni laboratories. The paper also describes a subjective validation test of the corpus, based on emotion-discrimination of two sentences carried out by two different groups of 24 listeners. The test was successful because it yielded an overall recognition accuracy of 80\%. It is observed that emotions less easy to recognize are joy and disgust, whereas the most easy to detect are anger, sadness and the neutral state.}
}

@inproceedings{dhall2014-EmotionRecognitionWild,
  title = {Emotion {{Recognition In The Wild Challenge}} 2014: {{Baseline}}, {{Data}} and {{Protocol}}},
  shorttitle = {Emotion {{Recognition In The Wild Challenge}} 2014},
  booktitle = {Proc. ICMI},
  author = {Dhall, Abhinav and Goecke, Roland and Joshi, Jyoti and Sikka, Karan and Gedeon, Tom},
  year = {2014},
  month = nov,
  %series = {{{ICMI}} '14},
  pages = {461--466},
  publisher = {Association for Computing Machinery},
  address = {New York, NY, USA},
  %%doi = {10.1145/2663204.2666275},
  %urldate = {2024-02-13},
  abstract = {The Second Emotion Recognition In The Wild Challenge (EmotiW) 2014 consists of an audio-video based emotion classification challenge, which mimics the real-world conditions. Traditionally, emotion recognition has been performed on data captured in constrained lab-controlled like environment. While this data was a good starting point, such lab controlled data poorly represents the environment and conditions faced in real-world situations. With the exponential increase in the number of video clips being uploaded online, it is worthwhile to explore the performance of emotion recognition methods that work `in the wild'. The goal of this Grand Challenge is to carry forward the common platform defined during EmotiW 2013, for evaluation of emotion recognition methods in real-world conditions. The database in the 2014 challenge is the Acted Facial Expression In Wild (AFEW) 4.0, which has been collected from movies showing close-to-real-world conditions. The paper describes the data partitions, the baseline method and the experimental protocol.},
  isbn = {978-1-4503-2885-2},
  keywords = {audio-video data corpus,emotion recognition in the wild,emotiw challenge}
}

@article{duville2021-MexicanEmotionalSpeech,
  title = {The {{Mexican Emotional Speech Database}} ({{MESD}}): Elaboration and Assessment Based on Machine Learning},
  shorttitle = {The {{Mexican Emotional Speech Database}} ({{MESD}})},
  author = {Duville, Mathilde M. and {Alonso-Valerdi}, Luz M. and {Ibarra-Zarate}, David I.},
  year = {2021},
  month = nov,
  journal = {Annual International Conference of the IEEE Engineering in Medicine and Biology Society. IEEE Engineering in Medicine and Biology Society. Annual International Conference},
  volume = {2021},
  pages = {1644--1647},
  %issn = {2694-0604},
  %doi = {10.1109/EMBC46164.2021.9629934},
  abstract = {The Mexican Emotional Speech Database is presented along with the evaluation of its reliability based on machine learning analysis. The database contains 864 voice recordings with six different prosodies: anger, disgust, fear, happiness, neutral, and sadness. Furthermore, three voice categories are included: female adult, male adult, and child. The following emotion recognition was reached for each category: 89.4\%, 93.9\% and 83.3\% accuracy on female, male and child voices, respectively.Clinical Relevance - Mexican Emotional Speech Database is a contribution to healthcare emotional speech data and can be used to help objective diagnosis and disease characterization.},
  langid = {english},
  pmid = {34891601},
  keywords = {Adult,Child,Emotions,Female,Humans,Machine Learning,Male,Reproducibility of Results,Speech,Voice}
}

@article{engberg1997-DesignRecordingVerification,
  title = {Design {{Recording}} and {{Verification}} of a {{Danish Emotional Speech Database}}: {{Design Recording}} and {{Verification}} of a {{Danish Emotional Speech Database}}},
  shorttitle = {Design {{Recording}} and {{Verification}} of a {{Danish Emotional Speech Database}}},
  author = {Engberg, Inger Sams{\o} and Hansen, Anya Varnich and Andersen, Ove Kjeld and Dalsgaard, Paul},
  year = {1997},
  journal = {EUROSPEECH'97 : 5th European Conference on Speech Communication and Technology, Patras, Rhodes, Greece, 22-25 September 1997},
  pages = {Vol. 4, pp. 1695-1698}
}

@inproceedings{hansen1997-GettingStartedSUSAS,
  title = {Getting Started with {{SUSAS}}: A Speech under Simulated and Actual Stress Database},
  shorttitle = {Getting Started with {{SUSAS}}},
  booktitle = {Proc. {{Eurospeech}} 1997},
  author = {Hansen, John H. L. and {Bou-Ghazale}, Sahar E.},
  year = {1997},
  pages = {1743--1746},
  %doi = {10.21437/Eurospeech.1997-494},
  %urldate = {2024-02-13}
}

@inproceedings{haq2009-SpeakerdependentAudiovisualEmotion,
  title = {Speaker-Dependent Audio-Visual Emotion Recognition},
  booktitle = {Proc. {{AVSP}} 2009},
  author = {Haq, Sanaul and Jackson, Philip J. B.},
  year = {2009},
  pages = {53--58},
  %urldate = {2024-02-13}
}

@article{lassalle2019-EUEmotionVoiceDatabase,
  title = {The {{EU-Emotion Voice Database}}},
  author = {Lassalle, Amandine and Pigat, Delia and O'Reilly, Helen and Berggen, Steve and {Fridenson-Hayo}, Shimrit and Tal, Shahar and Elfstr{\"o}m, Sigrid and R{\aa}de, Anna and Golan, Ofer and B{\"o}lte, Sven and {Baron-Cohen}, Simon and Lundqvist, Daniel},
  year = {2019},
  month = apr,
  journal = {Behavior Research Methods},
  volume = {51},
  number = {2},
  pages = {493--506},
  %issn = {1554-3528},
  %%doi = {10.3758/s13428-018-1048-1},
  %urldate = {2024-02-13},
  abstract = {In this study, we report the validation results of the EU-Emotion Voice Database, an emotional voice database available for scientific use, containing a total of 2,159 validated emotional voice stimuli. The EU-Emotion voice stimuli consist of audio-recordings of 54 actors, each uttering sentences with the intention of conveying 20 different emotional states (plus neutral). The database is organized in three separate emotional voice stimulus sets in three different languages (British English, Swedish, and Hebrew). These three sets were independently validated by large pools of participants in the UK, Sweden, and Israel. Participants' validation of the stimuli included emotion categorization accuracy and ratings of emotional valence, intensity, and arousal. Here we report the validation results for the emotional voice stimuli from each site and provide validation data to download as a supplement, so as to make these data available to the scientific community. The EU-Emotion Voice Database is part of the EU-Emotion Stimulus Set, which in addition contains stimuli of emotions expressed in the visual modality (by facial expression, body language, and social scene) and is freely available to use for academic research purposes.},
  langid = {english},
  keywords = {Emotion perception,Multisite validation,Voice stimuli set}
}

@misc{pichora-fuller2020-TorontoEmotionalSpeech,
  title = {Toronto Emotional Speech Set ({{TESS}})},
  author = {{Pichora-Fuller}, M. Kathleen and Dupuis, Kate},
  year = {2020},
  month = feb,
  publisher = {Borealis},
  %doi = {10.5683/SP2/E8H2MF},
  %urldate = {2024-02-13},
  abstract = {These stimuli were modeled on the Northwestern University Auditory Test No. 6 (NU-6; Tillman \& Carhart, 1966). A set of 200 target words were spoken in the carrier phrase "Say the word \_\_\_\_\_' by two actresses (aged 26 and 64 years) and recordings were made of the set portraying each of seven emotions (anger, disgust, fear, happiness, pleasant surprise, sadness, and neutral). There are 2800 stimuli in total. Two actresses were recruited from the Toronto area. Both actresses speak English as their first language, are university educated, and have musical training. Audiometric testing indicated that both actresses have thresholds within the normal range.},
  copyright = {http://creativecommons.org/licenses/by-nc/4.0},
  langid = {english},
  keywords = {Social Sciences}
}

@article{zhou2021emotional,
title = {Emotional voice conversion: Theory, databases and ESD},
journal = {Speech Communication},
volume = {137},
pages = {1-18},
year = {2022},
%issn = {0167-6393}
}

\end{document}


\maketitle
\begin{table*}[htbp]
\centering
\resizebox{1\textwidth}{!}{
\begin{tabular}{lrrrrrrrr}
\toprule
                            &   &   \multicolumn{6}{c}{\textbf{Performance in [\% UAR]}} \\
                            \cmidrule{2-9}
\multicolumn{1}{c}{\textbf{Datasets}} & \multicolumn{1}{c}{\textbf{W2V2}} & \multicolumn{1}{c}{\textbf{W2V2}} & \multicolumn{1}{c}{\textbf{Whisper}} & \multicolumn{1}{c}{\textbf{Whisper}} & \multicolumn{1}{c}{\textbf{HuBERT}} & \multicolumn{1}{c}{\textbf{HuBERT}} & \multicolumn{1}{c}{\textbf{HuBERT}} & \multicolumn{1}{c}{\textbf{HuBERT}} \\

\multicolumn{1}{c}{\textbf{}} & \multicolumn{1}{c}{\textbf{XLS-R}} & \multicolumn{1}{c}{\textbf{XLS-R}} & \multicolumn{1}{c}{\textbf{Medium}} & \multicolumn{1}{c}{\textbf{ Large }} & \multicolumn{1}{c}{\textbf{Large}} & \multicolumn{1}{c}{\textbf{ XLarge}} & \multicolumn{1}{c}{\textbf{Large}} & \multicolumn{1}{c}{\textbf{ Large }} \\

\multicolumn{1}{c}{\textbf{}} & \multicolumn{1}{c}{\textbf{300M}} & \multicolumn{1}{c}{\textbf{1B}} & \multicolumn{1}{c}{\textbf{}} & \multicolumn{1}{c}{\textbf{v3}} & \multicolumn{1}{c}{\textbf{}} & \multicolumn{1}{c}{\textbf{}} & \multicolumn{1}{c}{\textbf{\textsc{EmoSet} + 5}} & \multicolumn{1}{c}{\textbf{\textsc{EmoSet++}}} \\
\midrule    
Airplane Behaviour Corpus (ABC) \cite{schuller2007-AudiovisualBehaviorModeling}                 & 52.6 & 50.8 & 43.0 & 33.2 & 59.9 & 63.9 & \textbf{67.8} & 64.1 \\
Anger Detection (AD) \cite{Gerczuk22-EAT}                                                       & 89.3 & 87.7 & 63.4 & 75.7 & \textbf{90.4} & 89.7 & 89.2 & 83.2 \\
Burmese Emotional Speech (BES) \cite{nwe2003-SpeechEmotionRecognition}                          & 55.0 & \textbf{60.0} & 40.0 & 42.5 & 52.5 & 45.0 & 38.7 & 41.2 \\
CASIA \cite{-ChineseLDCOrgZiYuanChaKan}                                                         & 48.0 & 41.8 & 39.0 & 35.2 & 46.8 & \textbf{66.4} & 63.0 & 49.8 \\
Chinese Vocal Emotions (CVE) \cite{liu2012-RecognizingVocalEmotions}                            & 71.5 & 63.0 & 38.1 & 38.0 & 66.5 & 74.0 & 64.5 & \textbf{75.5} \\
Database of Elicited Mood in Speech (DEMOS) \cite{parada-cabaleiro2020-DEMoSItalianEmotional}   & 56.1 & 69.2 & 30.0 & 32.8 & 67.7 & 57.2 & 70.8 & \textbf{90.7} \\
Danish Emotional Speech (DES) \cite{engberg1997-DesignRecordingVerification}                    & 93.0 & 88.2 & 84.9 & 93.1 & 92.0 & \textbf{97.0} & 96.0 & 96.1 \\
EA-ACT \cite{schuller2006-AutomatischeEmotionserkennungAus}                                     & 73.0 & 85.0 & 39.0 & 39.0 & 63.0 & \textbf{85.0} & 56.0 & 70.0 \\
EA-BMW \cite{schuller2006-AutomatischeEmotionserkennungAus}                                     & 74.8 & 71.5 & 47.4 & 54.4 & 58.1 & \textbf{82.2} & 74.8 & 54.8 \\
EA-WSJ \cite{schuller2006-AutomatischeEmotionserkennungAus}                                     & 98.1 & 98.1 & 84.6 & 84.6 & 100 & 100 & 100 & \textbf{100} \\
Berlin Database of Emotional Speech (EMO-DB) \cite{burkhardt2005-DatabaseGermanEmotionala}      & 91.1 & 86.0 & 55.0 & 61.2 & 90.0 & 82.4 & 88.1 & \textbf{99.1} \\
eNTERFACE \cite{martin2006-ENTERFACE05AudioVisual}                                              & 66.4 & 76.1 & 39.3 & 38.6 & 93.9 & 63.2 & 92.9 & \textbf{94.6} \\
EU-Emotion Voice Database (EU-EV) \cite{lassalle2019-EUEmotionVoiceDatabase}                    & 36.3 & 42.3 & 37.1 & 33.0 & 41.1 & 29.8 & 41.4 & \textbf{92.8} \\
EmoFilm \cite{parada-cabaleiro2018-CategoricalVsDimensional}                                    & 56.9 & 43.4 & 49.7 & 44.5 & 58.6 & 57.4 & 58.9 & \textbf{60.5} \\
EmotiW-2014\cite{dhall2014-EmotionRecognitionWild}                                              & 40.0 & 34.7 & 28.6 & 27.2 & 33.1 & 32.5 & \textbf{40.2} & 39.1 \\
FAU Aibo \cite{batliner2008-ReleasingThoroughlyAnnotated}                                       & 36.2 & 35.3 & 31.1 & 25.8 & 38.2 & 31.5 & 39.1 & \textbf{39.8} \\
Geneva Emotion Portrayal  (GEMEP) \cite{scherer2010-BlueprintAffectiveComputing}                & \textbf{54.1} & 48.6 & 33.7 & 32.1 & 48.8 & 51.4 & 49.0 & 53.2 \\
Geneva Vocal Emotion Express (GVESS) \cite{banse1996-AcousticProfilesVocal}                     & \textbf{43.7} & 39.3 & 20.1 & 27.9 & 49.0 & 36.8 & 38.8 & 40.8 \\
IEMOCAP \cite{busso2008-IEMOCAPInteractiveEmotionalb}                                           & 56.4 & 49.7 & 42.9 & 39.0 & 61.1 & 65.8 & 63.9 & \textbf{67.8} \\
Multimodal EmotionLines Dataset (MELD) \cite{poria2019-MELDMultimodalMultiParty}                & 23.2 & 24.7 & 23.8 & 22.6 & 30.0 & 25.9 & 34.1 & \textbf{38.5} \\
Mandarin Emotional Speech (MES) \cite{nwe2003-SpeechEmotionRecognition}                         & 66.3 & 48.8 & 66.3 & 65.0 & 67.5 & 63.8 & \textbf{70.0} & 67.5 \\
PPMMK \cite{Gerczuk22-EAT}                                                                      & 45.0 & 46.1 & 33.6 & 31.1 & 51.1 & 55.9 & 51.1 & \textbf{61.4} \\
Speech in Minimal Invasive Surgery (SIMIS) \cite{schuller2010-SpeechMinimalInvasive}            & 26.3 & 25.3 & 25.6 & 25.3 & 26.4 & 26.0 & 29.6 & \textbf{32.1} \\
SmartKom Multimodal Corpus (SmartKom) \cite{schiel2002-SmartKomMultimodalCorpusb}               & 20.5 & 20.5 & 20.1 & 19.6 & 21.6 & 21.4 & 21.8 & \textbf{39.4} \\
Speech under Simulated and Actual Stress (SUSAS) \cite{hansen1997-GettingStartedSUSAS}          & 37.3 & 27.3 & \textbf{51.9} & 51.0 & 45.4 & 47.0 & 33.4 & 44.7 \\
TurkishEmo \cite{Gerczuk22-EAT}                                                                 & 63.6 & 58.0 & 46.6 & 46.6 & 72.7 & 73.9 & 68.2 & \textbf{80.7} \\
Toronto emotional speech set (TESS) \cite{pichora-fuller2020-TorontoEmotionalSpeech}            & -- & -- & -- & -- & -- & -- & 97.9 & \textbf{99.5} \\
EU-EmoSS \cite{oreilly2016-EUEmotionStimulusSet}                                                & -- & -- & -- & -- & -- & -- & -- & 85.1\\
Crowd-sourced Emotional Multimodal Actors Dataset (Crema-D) \cite{cao2014-CREMADCrowdsourcedEmotionala} & -- & -- & -- & -- & -- & -- & 68.8 & \textbf{79.9} \\
Ryerson Audio-Visual Database of Emotional Speech and Song (RAVDESS) \cite{livingstone2018-RyersonAudioVisualDatabasea} & -- & -- & -- & -- & -- & -- & 22.3 & \textbf{27.7} \\
Sharif Emotional Speech Database (ShEMO) \cite{mohamadnezami2019-ShEMOLargescaleValidated}      & -- & -- & -- & -- & -- & -- & 42.9	& \textbf{59.6}\\
Surrey Audio-Visual Expressed Emotion (SAVEE) \cite{haq2009-SpeakerdependentAudiovisualEmotion} & -- & -- & -- & -- & -- & -- & 51.8 & \textbf{63.4}\\
URDU \cite{latif2018-CrossLingualSpeech}                                                        & -- & -- & -- & -- & -- & -- & -- & 89.5\\
SUST Bangla Emotional Speech Corpus (SUBSECO) \cite{sultana2021-SUSTBanglaEmotional}            & -- & -- & -- & -- & -- & -- & -- & 61.7\\
Mexican Emotional Speech Database (MESD) \cite{duville2021-MexicanEmotionalSpeech}              & -- & -- & -- & -- & -- & -- & -- &  91.5\\
EMOVO \cite{costantini2014-EMOVOCorpusItalian}                                                  & -- & -- & -- & -- & -- & -- & -- & 65.1\\
Emotional Speech Dataset (ESD) \cite{zhou2021emotional,zhou2021-SeenUnseenEmotional}            & -- & -- & -- & -- & -- & -- & -- & 93.1\\

\bottomrule
\end{tabular}
}
\caption{Performance comparison of the applied Transformers on a wide range of speech emotion datasets. Our proposed fine-tuning of HuBERT Large on EMOSET++ demonstrates superior performance over all other Transformers.}
\label{tab:dataset_performance}
\end{table*}
\clearpage
\newpage

\section{\refname}
\printbibliography[heading=none]